\documentclass{article}
\usepackage{ijcai22}

\usepackage{times}
\usepackage{soul}
\usepackage{url}
\usepackage[hidelinks]{hyperref}
\usepackage[utf8]{inputenc}
\usepackage[small]{caption}
\usepackage{graphicx}
\usepackage{amsmath}
\usepackage{amsthm}
\usepackage{amssymb}
\usepackage{booktabs}
\usepackage{algorithm}
\usepackage{algorithmic}
\title{Investigating and Explaining the Frequency Bias in Image Classification}

\author{
Zhiyu Lin$^1$
\and
Yifei Gao$^1$\and
Jitao Sang$^{1,2}$\footnote{Corresponding Author}
\affiliations
$^1$Beijing Jiaotong University, China\\
$^2$Peng Cheng Lab, Shenzhen 518066, China\\
\emails
\{zyllin, yf-gao, jtsang\}@bjtu.edu.cn
}

\begin{document}

\maketitle

\begin{abstract}
    CNNs exhibit many behaviors different from humans, one of which is the capability of employing high-frequency components. This paper discusses the frequency bias phenomenon in image classification tasks: the high-frequency components are actually much less exploited than the low- and mid- frequency components. We first investigate the frequency bias phenomenon by presenting two observations on feature discrimination and learning priority. Furthermore, we hypothesize that (\romannumeral1) the spectral density, (\romannumeral2) class consistency directly affect the frequency bias. Specifically, our investigations verify that the spectral density of datasets mainly affects the learning priority, while the class consistency mainly affects the feature discrimination.
    \end{abstract}
    
    \section{Introduction}
    Convolutional neural networks (CNNs)  have now approached (and sometimes surpassed) "human-level" benchmarks on various tasks, especially those involving visual recognition. To understand this impressive performance, many recent findings show that DNNs differ in intriguing ways from human vision on processing the visual information~\cite{2019The}. One such human-model disparity is the texture bias in CNN-based image classification~\cite{Geirhos2019ImageNettrainedCA}. It has been observed that, unlike humans, CNNs tend to classify images by texture rather than by shape. Another interesting finding is that CNNs can exploit high-frequency image components that are not perceivable to human~\cite{2020High,2017Measuring}. The ability in capturing the high-frequency components of images partially explains the unintuitive behaviors of CNNs like generalization advancement and adversarial vulnerability.
    
\begin{figure*}
        \centering
        \includegraphics[scale=0.45]{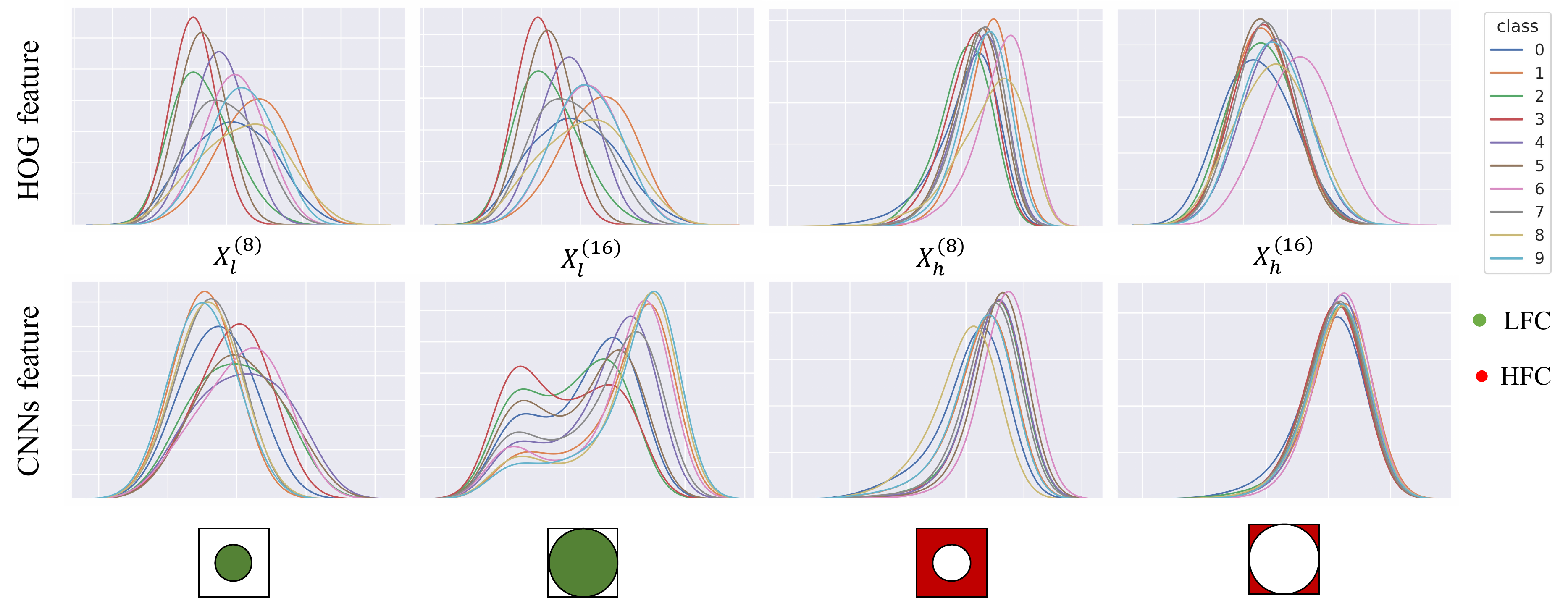}
        \caption{Visualization of KDE based inter-class variance for different frequency components (subfigures). We generate two types of frequency image following Eq.\eqref{eq:LandH}. The first row represents the class distribution of HOG features for all frequency components before CNNs extraction, it manifests noticeable discrimination between classes. The second row represents the features after CNNs extraction. While feature discrimination for the low- and middle- frequency components (left two sub-figures) are enhanced, the high-frequency components (right two sub-figures) are considerably inhibited.}
        \label{fig:figure1}
    \end{figure*}

    This paper follows the line of studies on analyzing CNNs' behavior in frequency domain. Our study is inspired by the observation that, although employed in classifying images, the high-frequency components are much less exploited than the low- and middle-frequency components. Fig.\ref{fig:figure1} illustrates the Kernel Density Estimation (KDE) curves of the 10 image classes in CIFAR-10 \cite{2009Learning} for low-, middle- and high-frequency components. It is shown that before CNNs feature extraction, HOG\cite{surasak2018histogram} feature for all frequency components manifest noticeable discrimination between classes. However, after CNNs feature extraction, while feature discrimination for the low- and middle-frequency components (left two sub-figures) are enhanced due to supervised learning, the high-frequency components (right two sub-figures) are considerably inhibited. Regarding the highest-frequency component in the rightmost sub-figure, the KDE curves of different classes almost collapse as one unique class. This demonstrates an obvious frequency bias phenomenon in image classification:  CNNs prefer low- and middle-frequency components over high-frequency components. Examining the frequency bias phenomenon and understanding the reasons behind will help fully exploit the potential of high-frequency components and further contribute to model improvement.
    
    Only a few studies have discussed the frequency bias phenomenon in image classification, with focus on exhibiting the biased accuracy by employing different frequency components. For example, \cite{2020High} discloses that low-frequency components are much more generalizable than high-frequency components, \cite{Abello_2021_CVPR} reports similar results by introducing a new way to divide frequency spectrum. In this paper, we investigate into more fundamental observations beyond the biased accuracy, and explain with novel hypothesis from data perspective what leads to the frequency bias\footnote{Our code is available at https://github.com/zhiyugege/FreqBias}. The contributions are summarized as two-fold: 
    
    \begin{itemize}
        \item We provide new observations along feature discrimination and learning priority to investigate the frequency bias phenomenon in terms of image classification tasks (Section 3). These two observations are correlated with each other, together offering supplementary perspectives to recognized biased accuracy in existing studies.
               
        \item We propose hypotheses to explain the frequency bias from perspective of spectral density and class consistency (Section 4). Experiment results and analyses verify our hypotheses, which sheds light on future solution to alleviate the frequency bias phenomenon.
    
    \end{itemize}

    %% 这里开始写文章的公式
    \section{Notations and Preliminaries}\label{Sec:Notions}
    In this section, we set up the basic notations used in this paper. Given a data sample $(X, y)$ where $X$ represents the image with corresponding label $y$, $f(X; \theta) \in \mathbb{R}^{D}$  denotes the feature of a specific intermediate layer of the CNNs, $D$ denotes the dimension of feature. Unless otherwise specified, we use the outputs from the penultimate layer as the feature representation for further analysis. $\hat{y} = g(f(X; \theta), w) \in \mathbb{R}^{C}$ denotes the output of the CNNs, where $C$ is the number of classes. $\mathcal{L}(y, \hat{y})$ denotes a generic loss function (e.g., cross-entropy loss). In this study, we conduct frequency analysis with the 2D-discrete Fourier transform (shorten as DFT) computed as follows:\footnote{$H, W$ denotes the height and width of input image.}
    \begin{equation}\label{eq:DFT}
        \mathcal{F}(X)(u, v) = \dfrac{1}{HW}\sum\limits_{h=0}^{H-1}\sum\limits_{w=0}^{W-1} X(h, w)\cdot e^{-j2\pi (\frac{uh}{H}+\frac{vw}{W})}
    \end{equation}
    where $\mathcal{F}(\cdot)$ denotes DFT and $X_{\mathcal{F}} = \mathcal{F}(X)$ is the transformed Fourier image in frequency domain with the same dimension as the input image. $\mathcal{F}^{-1}(\cdot)$ denote the 2D-inverse discrete Fourier transform (shorten as IDFT). We decompose the original image, $ X = \{X_{l},X_{h}\}$, where $X_{l}$ and $X_{h}$ denote the low-frequency components (shortened as LFC) and high-frequency components (shortened as HFC)\footnote{Similarly, MFC indicates the mid-frequency components.}. Thus, we have the following two equations:
    \begin{equation}\label{eq:LandH} 
        \left\{  
            \begin{array}{cc}
                X_{l}^{(r)} = \mathcal{F}^{-1}(\mathcal{F}(X)\odot \mathcal{M}_{l}^{(r)}) & \\
                \\
                X_{h}^{(r)} = \mathcal{F}^{-1}(\mathcal{F}(X)\odot \mathcal{M}_{h}^{(r)})  &\\
            \end{array}
        \right.  
    \end{equation}
    where $\mathcal{M}^{(r)}$ denotes the matrix of characteristic function with radius $r$, $\odot$ denotes the Hadamard product. We consider $d(\cdot,\cdot)$ as the  Euclidean distance when dividing the spectrum in bands, and $(c_{i},c_{j})$ as the center of an image. Therefore, we have: 
    \begin{align} \label{eq:Mask Matrix}
        \mathcal{M}_{l}^{(r)}(i, j) &= \left\{
            \begin{array}{cc}
                1, & \quad  if \;\; d((i,j), (c_{i}, c_{j})) \leq r \\  
                \\
                0, & \quad otherwise 
            \end{array}
        \right.                      
    \end{align}
    Obviously, there is $\mathcal{M}_{h}^{(r)} = 1-\mathcal{M}_{l}^{(r)}$.
    \vspace{0.1cm}
    
    \noindent We follow the convention in \cite{2020Watch} and compute the azimuthally average of the magnitude of Fourier coefficients over radial frequencies to obtain the reduced spectrum and normalize it. Specifically, the spectral density is defined as the azimuthal integration over radial frequencies $\phi$.
    \begin{equation}\label{eq:Spectral Density}
        AI_{k}(X) = \int_{0}^{2\pi} \Vert \mathcal{F}(X)(\omega_{k}\cdot cos\phi,\omega_{k}\cdot sin\phi) \Vert ^{2} d\phi 
    \end{equation}
    where $k=0,\cdots H/2-1$.

    %% 这里切入第一个bias现象
    \section{Observations of Frequency Bias}
    
    \subsection{On Feature Discrimination }
    Following the results in Fig.~\ref{fig:figure1} that the discriminative capability of HFC is inhibited after CNNs feature extraction, in this subsection, we investigate the frequency bias phenomenon by examining feature discrimination among different frequency components. Specifically, we propose to measure feature discrimination by computing the inter-class variance on KDE-based class feature distribution.
    
    \subsubsection{KDE-based Inter-Class Variance Calculation}
\begin{figure}
        \centering
        \includegraphics[scale=0.24]{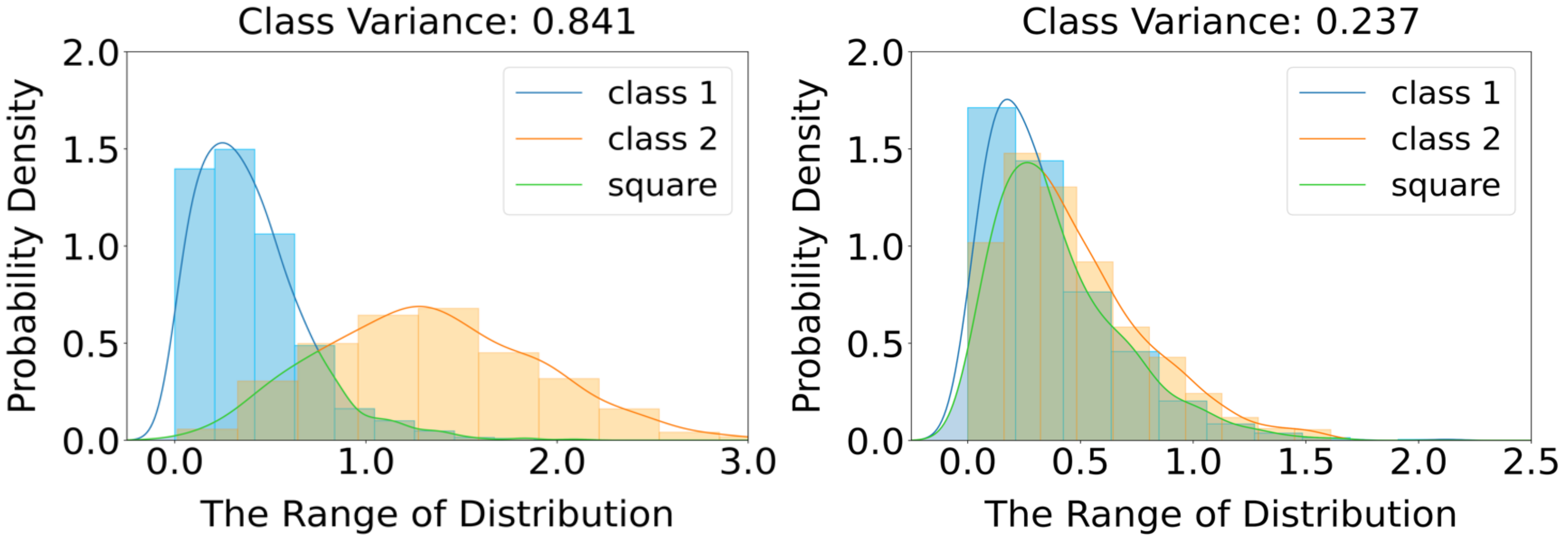}
        \caption{Illustration of KDE-based inter-class variance calculation for two feature dimensions (subfigure). For each feature dimension, the inter-class variance is estimated by the overlapped ratio (green area)  of the two class curves (orange and blue).}
        \label{fig:class-variance}
    \vspace{-0.1cm}
    \end{figure}

    We first introduce an estimation of class feature distribution based on KDE. Regarding the $\mathnormal{k^{th}}$ feature dimension of class $c$, we sample $N$ feature points $\{x_{i} \}_{i=1}^{N}$ and estimate the kernel density on feature value $x$ as follows:
    \begin{equation}\label{eq:kde}
        S_{c}^k(x) = \dfrac{1}{N}\sum\limits_{i=1}^{N}K(x-x_{i}) = \dfrac{1}{Nh}\sum\limits_{i=1}^{N}K(\dfrac{x-x_{i}}{h})
    \end{equation}
    where $\mathnormal{K(\cdot)}$ denotes the Gaussian kernel function and $\mathnormal{h}$ is the bandwidth selected by Silverman's rule. Enumerating all possible feature values derives the feature distribution vector $\mathbf{S}_{c}^k$, which is illustrated by the discrete histograms under each curve in Fig.\ref{fig:class-variance}. The set of feature distributions $\{S_{c}^{k} \}_{k=1}^{D}$ over all feature dimensions can be viewed as an estimation of the class feature distribution. Note that the curve of each class distribution in Fig.\ref{fig:figure1} is estimated by all feature dimensions.
    
    To evaluate the inter-class variance between two classes $c_i$ and $c_j$, we propose to compute the overlapped area ratio between their feature distribution curves. Specifically, the overlapped area ratio is first computed for each feature dimension $k$ and then averaged over all the $D$ dimensions. Formally, it is defined as:
    \begin{equation}
        \mbox{Inter}(c_{i}, c_{j}) = \dfrac{1}{D}\sum\limits_{k=1}^{D} \sum_x \dfrac {\min(S_{c_i}^k(x),S_{c_j}^k(x))} {\max(S_{c_i}^k(x),S_{c_j}^k(x))}
    \end{equation}
    where $\min(\cdot)$,  $\max(\cdot)$ are minimization and maximization functions respectively. Finally we measure the over inter-class variance by averaging all class pairs: 
    \begin{equation}
        \mbox{Variance} = 1 - \dfrac{2}{C(C-1)} \sum\limits_{i=1}^{C} \sum\limits_{j=i}^{C} \mbox{Inter}(c_{i}, c_{j})
    \end{equation}
    %% 插入单个维度的差异图像
    
    \subsubsection{Observations and Discussions}
\begin{figure}
        \centering
        \includegraphics[scale=0.25]{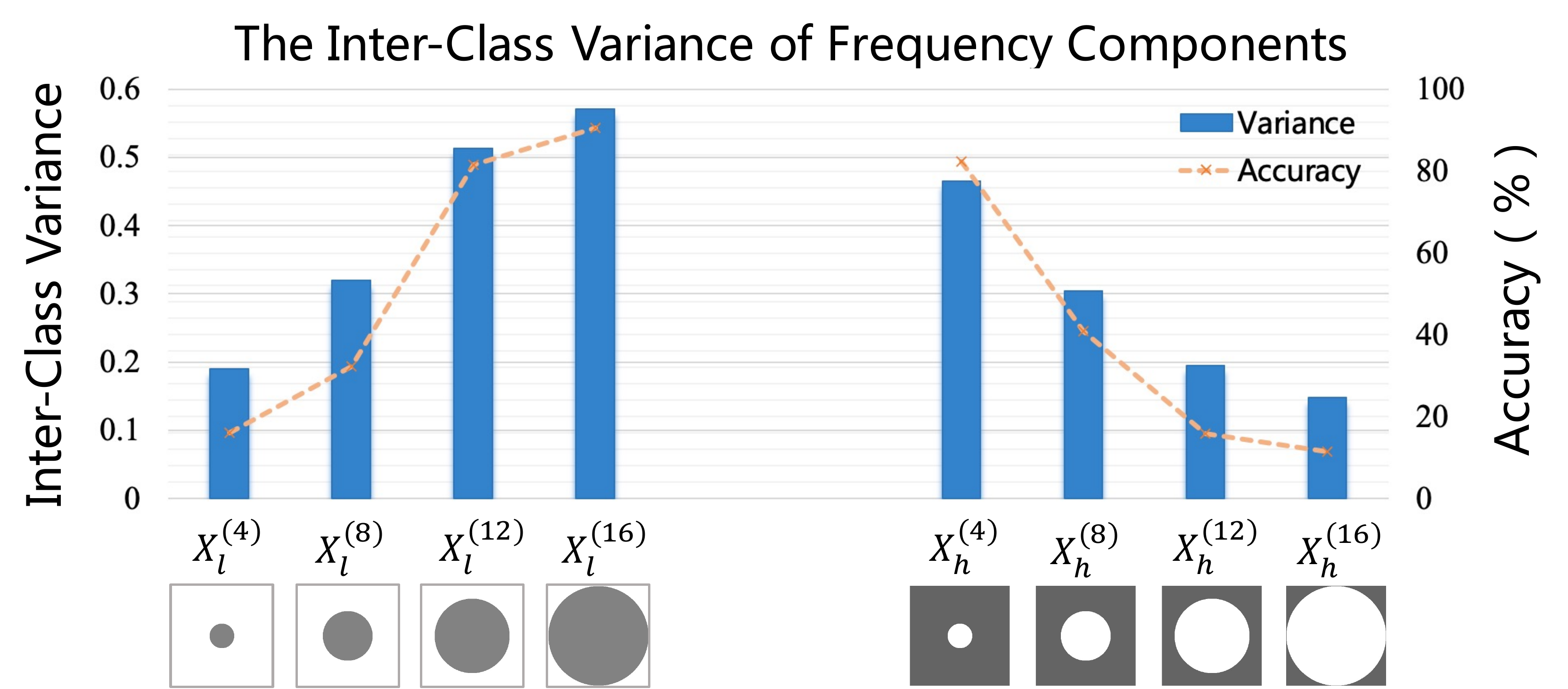}
        \caption{Results of KDE-based inter-class variance and test accuracy between different frequency components. A positive correlation exhibits between variance and test accuracy.}
        \label{fig:sec3}
    \end{figure}

    We report results with ResNet-50 \cite{2016Deep} trained on CIFAR10 dataset~\footnote{All experiments are repeated using ResNet-18 model and Restricted ImageNet dataset \cite{2009ImageNet}. The key observations are consistent. Detailed results are available in appendix.}. To investigate the feature discrimination for different frequency components, for LFC and HFC, we respectively select frequency components with $\mathnormal{r}=\{4,8,12,16\}$ (defined as Eq.\eqref{eq:LandH}). For each frequency component, the inter-class variance and test accuracy are calculated and shown in Fig.\ref{fig:sec3}.
    
    The first observation is that feature discrimination capability exhibits significant bias among different frequency components. While features corresponding to the LMFC (e.g., $X_{l}^{(12)}$, $X_{l}^{(16)}$, $X_{h}^{(4)}$) capture adequate inter-class variance, the high- (e.g., $X_{h}^{(12)}$, $X_{h}^{(16)}$)  and very low- (e.g., $X_{l}^{(4)}$, $X_{l}^{(8)}$) frequency components contribute to trivial discriminative features. With the fact that the inter-class variance reduces rapidly for the HFC as radius $r$ increases, we demonstrate CNN's frequency bias from the perspective of feature discrimination that higher-frequency components are less exploited. A second observation is on the positive correlation between inter-class variance and test accuracy. It is easy to understand that discriminative features will contribute more to model's generalization performance. This endorses the observed accuracy bias among frequency components in previous studies.

    %~\footnote{Note that CNN also shows bias on the very low-frequency components w.r.t. feature discrimination. In this paper, we focus on discussing the bias phenomenon on the high frequency components. We exclude the very low-frequency parts when mentioning the low-frequency components (LFC).}
    
    %% 这里切入第二个bias现象
    \subsection{On Learning Priority}
    The previous analysis validates the phenomenon of feature discriminative bias led by frequency bias, which can be viewed as an explanation for generalization bias observed in previous works \cite{2020High}. This intuitively leads to an  question: \emph{How frequency bias affects the learning process?} While the affect of frequency bias can be effectively observed after training, understanding the evolution of bias phenomenon during training remains a problem. In this subsection, we further observe the learning priority of frequency components during training to investigate frequency bias phenomenon. We propose to evaluate the learning priority by analyzing gradient information.
    
    \subsubsection{Gradient Evaluation of Frequency Components}
    In the case of classification task, the measurement of gradients $\frac{\partial{\mathcal{L}}}{\partial{X}}$ provides us valuable information about the contribution of spatial domain image $X$ to loss function. In our work, we pay more attention on the gradient information of frequency domain image. Therefore, we first introduce a gradient map as an evaluation metric to measure the learning priority from frequency perspective. It is formally defined as: $\frac{\partial{\mathcal{L}}}{\partial{X^{(k)}}}$, where $X_{k}$ represents a spatial image with only $k^{th}$ frequency band information reserved from original image $X$. We expect to explore the priority performance of each frequency band of an image. One straight way to get the complete gradient information is as follows: we should first split the image spectrum into $N$ bands, then traverse the value of $k$ and use IDFT to obtain spatial images, finally compute the gradients for $N$ times. However, directly calculating $\frac{\partial{\mathcal{L}}}{\partial{X^{(k)}}}$  involves a high computational cost, we introduce the following proposition for approximated calculation. The proof of the proposition is available in appendix. 
    
    \vspace{0.4cm}
    %% 这里给出学习优先级的命题
    \noindent \textbf{Proposition} \emph{$X$ denotes an input image. $\mathcal{M}^{(k)}$ is the  matrix of characteristic function which preserve the $k^{th}$ band. Then the gradient map can be represented by the following equation: $\dfrac{\partial\mathcal{L}}{\partial X^{(k)}} = \mathcal{F}^{-1}(\mathcal{F}(\dfrac{\partial\mathcal{L}}{\partial X}) \odot \mathcal{M}^{(k)})$, where $\mathcal{F}(\dfrac{\partial\mathcal{L}}{\partial X})$ denotes the gradient spectrum.}
    \vspace{0.4cm}
    
    %% 这里给出可视化学习优先级的算法
    \begin{algorithm}[tb]
        \caption{Visualizing the learning priority}
        \label{alg:algorithm}
        \textbf{Input}: $X$: test image, $N$: training epochs\\
        \textbf{Parameter}: The model parameters of each epoch\\
        \textbf{Output}: The gradient spectrum
        \begin{algorithmic}[1] %[1] enables line numbers
        \FOR{ epoch $\leftarrow$ 0 to $N$}
        \STATE $\mathcal{L} \leftarrow$ Compute the loss
        \STATE $\mathcal{G} \leftarrow$ Loss backward and obtain the gradient map
        \STATE $\mathcal{S} \leftarrow AI(\mathcal{G})$ Compute the spectral density of gradient
        \ENDFOR
        \STATE \textbf{return} spectral density of gradients $\mathcal{S}$
        \end{algorithmic}
    \end{algorithm}

    Following the proposition, the above computation process is equivalent to preserve the $k^{th}$ band of the gradient spectrum in practice. It substantially reduces the computational cost, where we just need to compute the gradients once. In order to visualize the evolution of learning priority, we calculate the spectral density of gradients in each training epoch. The details are shown in Algo.\ref{alg:algorithm}.
    
    \subsubsection{Results and Discussions}
\begin{figure}
    \centering
    \includegraphics[scale=0.22]{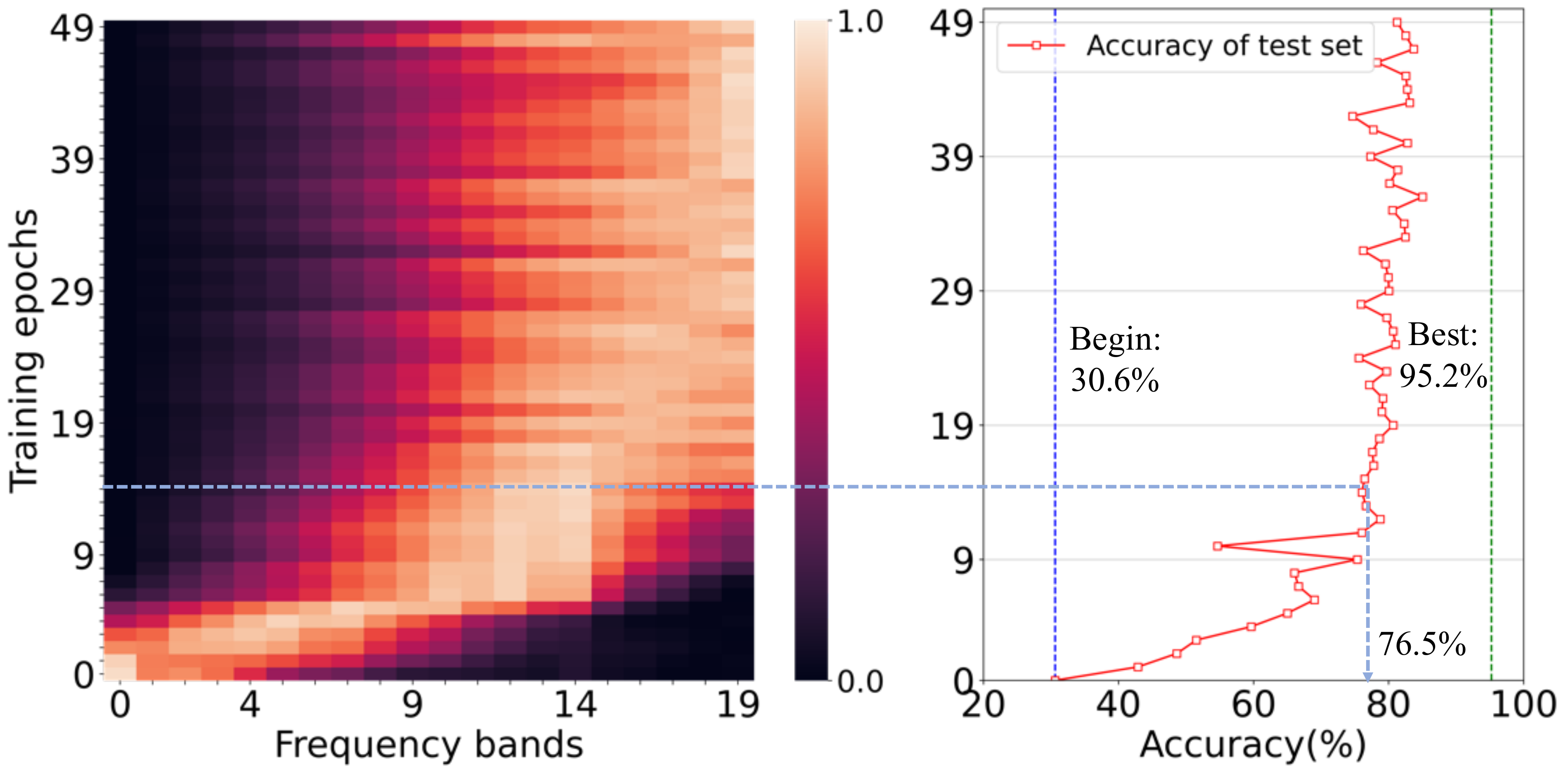}
    \caption{(Left): Evolution of the learned frequency components (x-axis for frequency bands) during the first 50 training epochs (y-axis). The colors show the measured amplitude of the gradients at the corresponding frequency bands (normalized to $[0,1]$). (Right): Evolution of the testing accuracy(x-axis for value of accuracy) during the first 50 training epochs (y-axis).$\;$ The peaks of the gradient density is related to the learning priority, suggesting that models are strongly biased towards learning the LFC. Also, the LFC contribute more than HFC to accuracy.}
    \label{fig:lp-with-loss}
    \end{figure}

    \vspace{-0.1cm}
    The training setting is same as mentioned in section 3.1. To analyze the evolution of learning priority, we compute spectral density of gradients and average it over test set. We exhibit the evolution results during the first 50 epochs in Fig.\ref{fig:lp-with-loss} (left). The spectral density of gradients at each row is notable: it mainly concentrates on some specific frequency bands (i.e., the colors on some frequency bands are much brighter than others) rather than uniformly distributes in all frequency bands, suggesting that the model exclusively pays attention to learning specific frequency information at each training stage. By comparing different rows, we find that the peak of gradient density (i.e., the brightest color) centralizes in the low-frequency bands at the early training stage, and then gradually shifts towards the middle-frequency bands and finally stop at the high-frequency bands. This significant trend, which reflects that models are strongly biased towards learning LFC first, reveals the learning priority bias on different components.
    
    Interestingly, we find that the learning priority shifting from low frequency to high frequency occurs in a short time (about 20 epochs). Meanwhile, model focuses on learning HFC in the rest of training process (about 180 epochs). Recalling the discussion in Section 3.1, we illustrate accuracy curve of test set in Fig.\ref{fig:lp-with-loss} (right). We mark the position of $15^{th}$ epoch with a dash line. Note that test accuracy during the first 15 epochs increases much faster than the rest of training epochs. We argue that the bias of learning priority towards LFC contributes much to the generalizing capacity of model. Inversely, while the model manages to pick up HFC in a long training stage, it eventually achieves a poor performance on generalization improvement.

    %% 这里开始写第四节的Toy实验
    \section{Hypothesis behind Frequency Bias}
    
    \subsection{On Spectral Density}
    
    To explain the frequency bias, we start with the investigation of  frequency-domain characteristics on spectral density. Recent studies \cite{2021On} show that the architectures of GANs exhibit a frequency bias phenomenon towards generation task. They reports that an enhanced density of high frequency in spectrum is beneficial for the reconstruction of HFC. Motivated by this observation, we hypothesize that spectral density can serve as an explanation for frequency bias in terms of image classification. In this subsection, we propose a framework called \emph{Convolutional Density Enhancement Strategy} (CDES) to modify spectral density of natural image and observe the performance changes in feature discrimination and learning priority.
    
    \subsubsection{Convolutional Density Enhancement Strategy (CDES)}
    
    Following the spectrum settings in \cite{2021On}, the main idea of our framework is to create a density peak in high frequency. Since simply adding noise is difficult to modify the spectral density as we expect, we first propose to perform convolution operations on original images with a set of trainable convolution filters. Inspired by \cite{2020Watch,2019Fourier}, the filters are optimized with loss function defined as:
    \vspace{-0.2cm}
    \begin{equation}\label{eq:Spectrum Loss}
        \begin{aligned}
            \mathcal{L} = \sum\limits_{k=0}^{H/2-1} \Vert AI_{k}(X^{(Conv)}) - AI_{k}^{(Target)}\Vert_{2}^{2} \\ +\Vert X^{(Conv)}-X \Vert_{2}^{2}
        \end{aligned}
    \end{equation}
    where $X^{(Conv)}$ denotes the images after convolution operation, $AI^{(Target)}$ denotes the target spectral density. The first term of Eq.\eqref{eq:Spectrum Loss} is to match the spectral density of $X^{(Conv)}$ with target spectral density $AI^{(Target)}$ with L2-loss. To ensure the semantic information of the image, a regularization constraint at pixel level is illustrated in the second term of Eq.\eqref{eq:Spectrum Loss}. Since we are more concerned about the bias phenomenon towards HFC, we aim to maintain the immutability on LFC. Considering the disturbance on LFC after the above operations, we replace the LFC of $X^{(Conv)}$ with LFC from original image $X$ under a specified radius $\mathnormal{r}$. We select the radius of lowest point in expected spectral density curve. Then the recombined image $X^{(Rec)}$ is defined as the following equation.
\begin{equation}\label{eq:Recombine Image}
    \resizebox{.9\linewidth}{!}{$
            X^{(Rec)} = \mathcal{F}^{-1}(\mathcal{F}(X) \odot\mathcal{M}_{l}^{(r)} + \mathcal{F}(X^{(Conv)}) \odot \mathcal{M}_{h}^{(r)})
        $}
\end{equation}
\vspace{-0.3cm}
    
    \noindent The new dataset thus consists of images $X^{(Rec)}$ and we split the train and test set on the new datasets.
    
    \begin{table}
        \centering
        \begin{tabular}{ccccccc}
            \toprule 
            Acc(\%)  & $X_{h}^{(10)}$ & $X_{h}^{(12)}$ & $X_{h}^{(14)}$ & $X_{h}^{(16)}$ & $X_{h}^{(18)}$\\
            \midrule
            $f_{scr}$    & \textbf{79.0}  & \textbf{68.2}  & \textbf{42.4} & \textbf{30.4} & \textbf{30.9}     \\
            $f_{wcr}$    & 20.4 & 15.5& 12.8& 10.0& 10.0     \\
            $f_{baseline}$ & 43.2 & 27.2& 14.3& 10.5& 8.3  \\
            \bottomrule
        \end{tabular}
        \caption{Here we report the test accuracy of the HFC among three models, $f_{scr}$ significantly improves the generalization of HFC, while $f_{wcr}$  has a poor performance that is even worse than the $f_{baseline}$.}
        \label{tab: table1}
    \end{table}

    \subsubsection{Experiment Setting}
    
    In our settings, we construct two variant datasets of CIFAR-10 based on CDES. (\romannumeral1) $\mathcal{D}_{wcr}$: We train one group of filters for the dataset. Note that this definition ensures that the filters is indistinguishable between classes. We argue that the added HFC is weakly class-related. (\romannumeral2) $\mathcal{D}_{scr}$: We separately train a group of filters corresponding to each class, where the added HFC are strongly class-related\footnote{It’s worth noting that the expected spectrum of two datasets are the same and we respectively split the train and test set.}. Given the  $\mathcal{D}_{wcr}$, $\mathcal{D}_{scr}$ and original datasets, we train the following classifiers: $f_{wcr}$,$f_{scr}$, $f_{baseline}$, and test them on the corresponding test set. The settings of the rest paper is to run the experiment with 150 epochs with SGD Momentum optimizer with learning rate set to be 1e-2 and batch size set to be 100. All experiments are repeated three times and averaged.
    
    \vspace{0.2cm}
    \noindent \paragraph{Analyses of Feature Discrimination.} We report the test accuracy of HFC in Tab.\ref{tab: table1} with the following observations: (\romannumeral1) $f_{scr}$ improves the feature discrimination on HFC with significantly increased accuracy. (\romannumeral2) $f_{wcr}$ exhibits a even worse performance of feature discrimination than $f_{baseline}$. We owe this huge disparity to the strategy of class-specific filters since the added HFC is weakly class-related. This also indicates that spectral density is not the only factor to explain the feature discrimination performance. 
    
    \vspace{0.2cm}

    \begin{figure}
    \centering
    \includegraphics[scale=0.27]{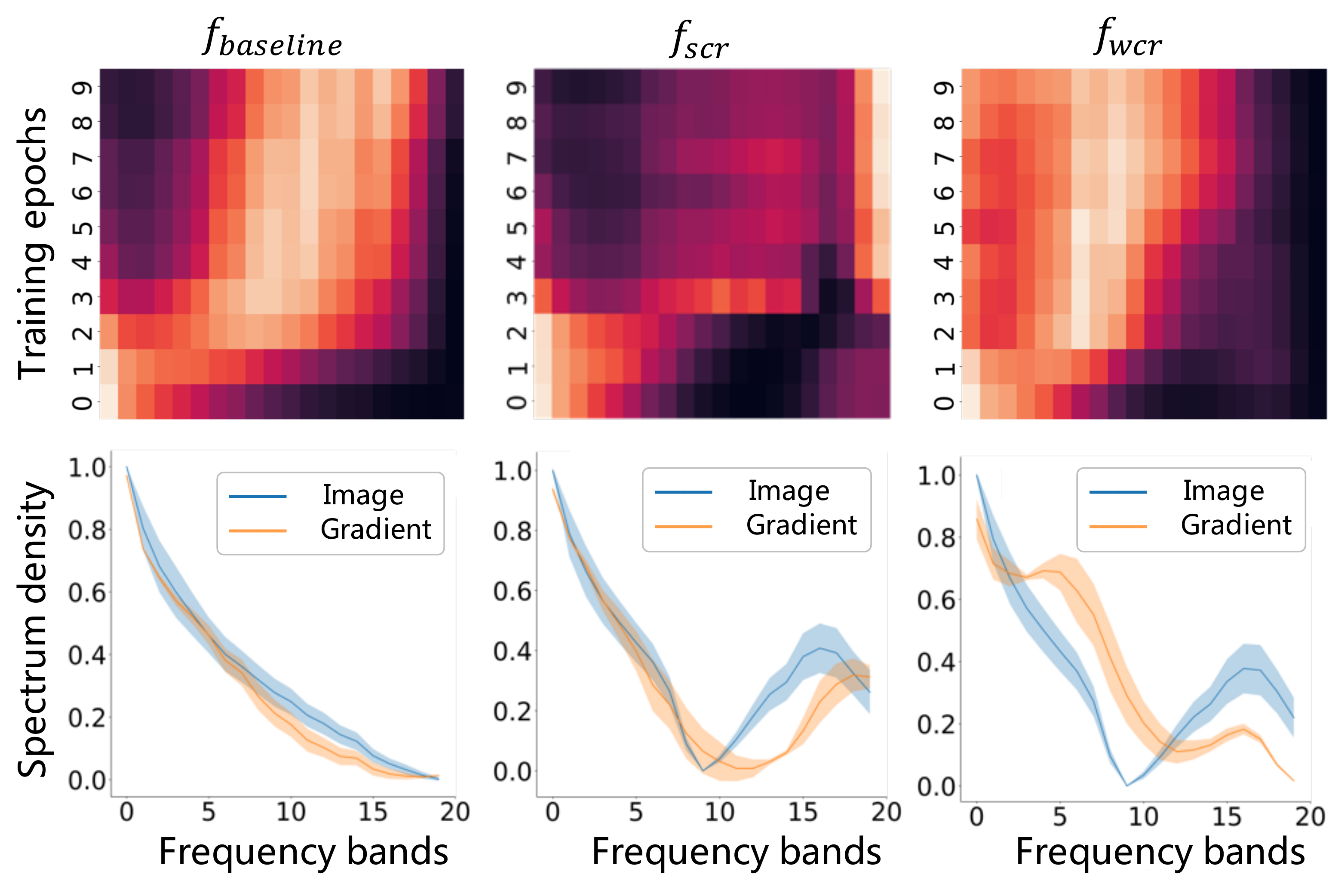}
    \caption{The first row shows the evolution of gradient spectrum in the first 10 training epochs among three models. The second row shows the Comparison of gradients and corresponding image on spectral density. The similar trends of spectral density at the early training stage suggests that the spectral density of dataset leads to the learning priority.}
    \label{fig:lp-match}
    \end{figure}
\vspace{-0.3cm}

    \noindent \paragraph{Analyses of Learning Priority.} We visualize the gradient evolution during the first 10 epochs in the first row of Fig.\ref{fig:lp-match}. An interesting finding is that the evolution tendency of $f_{scr}$ and $f_{wcr}$  are similar in the first few epochs. We hypothesize that the consistent spectral density of $\mathcal{D}_{wcr}$ and $\mathcal{D}_{scr}$ leads to the similar tendency. To verify this hypothesis, we average the spectral density of gradients in first 3 epochs and compare it with the spectral density of datasets. The second row of Fig.\ref{fig:lp-match} confirms our hypothesis with the similar trends of two spectral density curves. This inspires us that the learning priority is led by the spectral density of the dataset at the early training stage.
    
    The observations and analyses  above partially verify our hypothesis that the spectral density of datasets explains well on learning priority phenomenon but is not a sufficient condition for feature discrimination.

    %% 类一致性表达
    \subsection{On Class Consistency}
    In typical classification problem, the classifier is optimized to learn a consistent feature representation within the same class, so as to model the data-label relationship existed in the dataset. In this subsection, we aim to explain the frequency bias from the perspective of class consistency. Since data augmentation is an important training trick to ensure the class consistency at the data side, we first try to understand its validity and outstanding performance from frequency perspective. We choose the type of Mixed Sample Data Augmentation (shorten as MSDA) as our analyzing target.The method so far proposed can be broadly categorized as either combining samples with interpolation (e.g., Mixup\cite{zhang2017mixup} and CutMix\cite{yun2019cutmix}).

    \subsubsection{Frequency-based MSDA Analyses}
\begin{figure}[ht]
    \centering
    \includegraphics[scale=0.28]{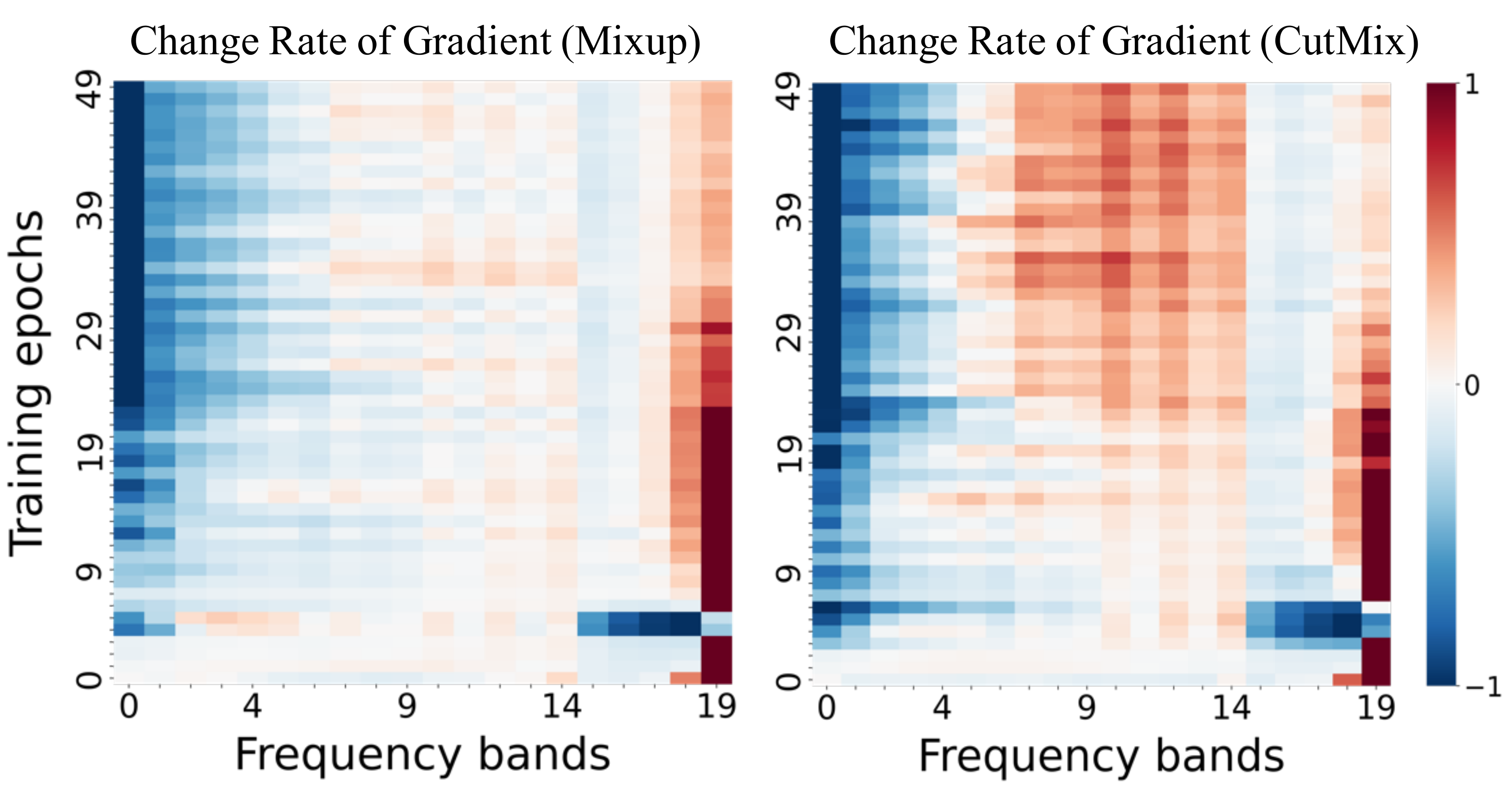}
    \caption{Visualization on the change rate of gradient evolution. Left reports the Mixup model, Right reports the CutMix model. Both of them show a tendency of utilizing MFC when the training process tends to be stable.}
    \label{fig:MSDA}
    \vspace{-0.2cm}
    \end{figure}

    To test the impact of MSDA strategy on the learning priority of  different frequency bands, we use  
    $$ \dfrac{AI_{k}{(G^{(MSDA)})}-AI_{k}(G)}{AI_{k}(G)}$$ (for $k=1,2, H/2-1$, where $G^{(MSDA)}$ and $G$ denotes the gradient map of model trained with and without MSDA respectively) as a metric to represent the change rate of learning priority. We normalize the value to [-1,1] and results are shown in Fig.\ref{fig:MSDA}.
    
    An overall observation shows that MSDA models increase the utilization of MFC. We notice a significantly increasing trend of the spectral density of gradient on MFC when the training process tends to be stable. The generalization results\footnote{The generalization results are available in appendix.} of MFC also report a better performance on MSDA model than the baseline model (i.e., when $r=8$ or $r=12$), which align well with the phenomenon of learning priority. These findings imply that MSDA preserve the class consistency on MFC, which can effectively boost the performance of classification.

    \subsubsection{Frequency Recombination-based Data Augmentation}
    
    Justified by the above analysis and inspired by the strategy of MSDA, we expect to introduce the idea of class consistency to the frequency domain. We hypothesize that the model only pick up the frequency components with the property of class consistency. For example, if we break up the class consistency on LFC, model will pay more attention to employing HFC.
    
    In order to verify this hypothesis, we construct another variant dataset of CIFAR-10. We modify each image-label pair as follows: Given a clean image $\mathnormal{X}$ with label $\mathnormal{y}$, we uniformly select another image $\mathnormal{X'}$ with label $\mathnormal{y'}$ from the dataset, (i.e.,$\mathnormal{y}\neq\mathnormal{y'}$,). Then we recombine the LFC of $\mathnormal{X'}$ (shorten as $x'_{l}$) and the HFC of $\mathnormal{X}$ (shorten as $X_{h}$) to create the new image $\mathnormal{X^{*}}$. For the selection strategy of radius $\mathnormal{r}$ in Eq.\eqref{eq:Mask Matrix}, we consider the following two constraints: (\romannumeral1) For $X'_{l}$, it is indistinguishable from $\mathnormal{X'}$ to human observation. (\romannumeral2) For $X_{h}$, it preserves high frequency information from $X$ as much as possible. In this case, we label $\mathnormal{X^*}$ with label $\mathnormal{y}$ so as to break the class consistency of LFC established by human annotation and preserve the class consistency of HFC. Then we create a new dataset named with HARS-dataset with image-label pair as ($\mathnormal{X^*}$,$\mathnormal{y}$).
    
\begin{figure}[]
    \centering
    \includegraphics[scale=0.24]{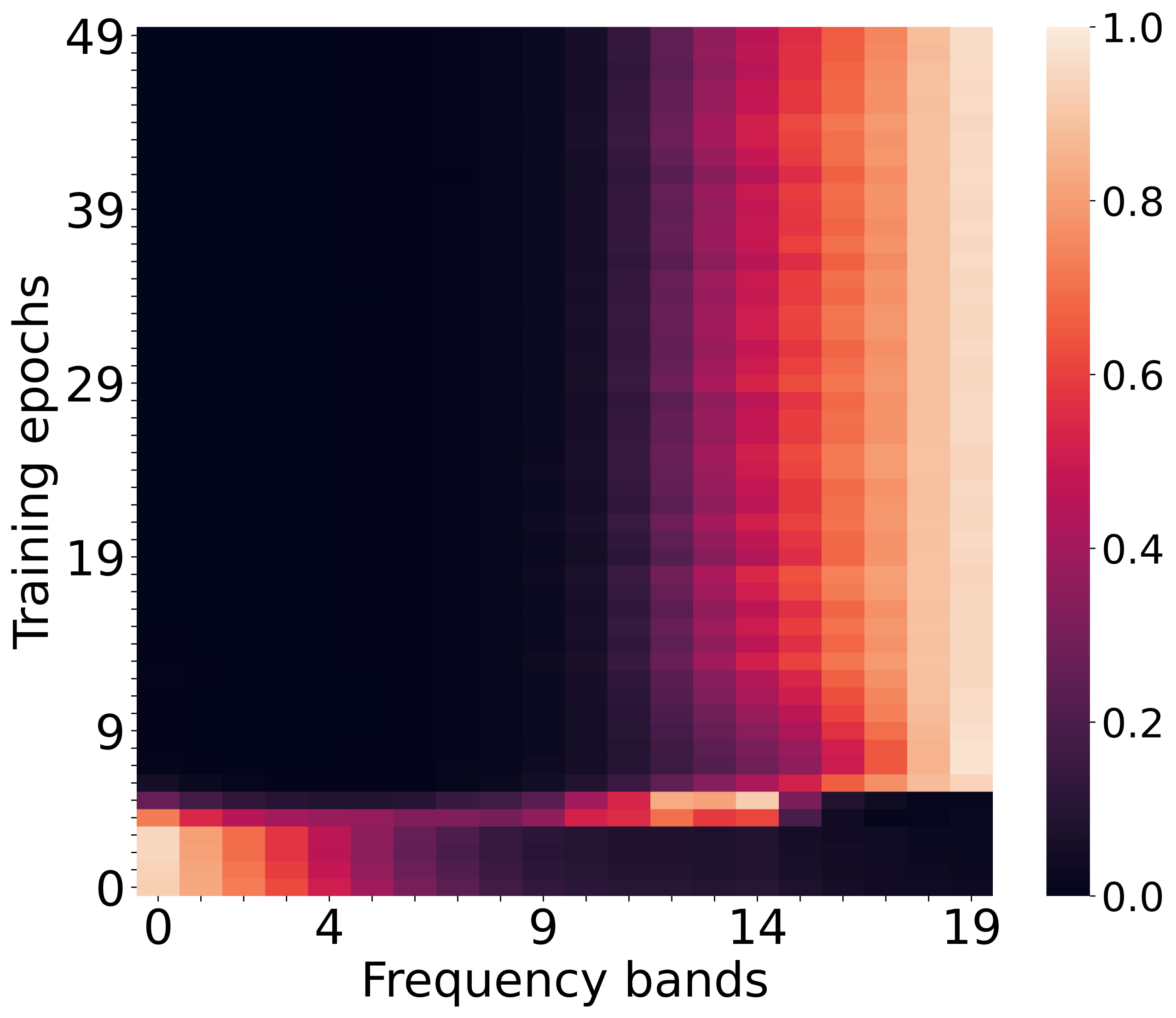}
    \caption{We visualize the evolution of gradient spectrum of HARS model in the first 50 epochs.}
    \label{fig: HF-restruct}
    \end{figure}

    We report the generalization performance in Tab.\ref{tab: table2} and the results of learning priority are plotted in Fig.\ref{fig: HF-restruct}. We have the following interesting findings:
    
    \vspace{0.1cm}
    \noindent \paragraph{Analyses of Feature Discrimination.} The generalization performance of HFC on the clean test set is significantly enhanced. This indicates that training on HARS-dataset effectively increase the employment of HFC as well as the discrimination of high frequency features. 
    
    \vspace{0.1cm}
    \noindent \paragraph{Analyses of Learning Priority.} (\romannumeral1) Although we have broken up the class consistency of LFC, model still picked up LFC at the early training stage. The reason for the remaining bias towards LFC is due to the dominate power of LFC’s density in spectrum, which is consistent with original dataset. This observation further justify our hypothesis that the spectral density affects learning priority of frequency information. (\romannumeral2) Compared with the gradient spectrum evolution of vanilla set-up as shown in Fig.\ref{fig: HF-restruct}, we find that model starts paying attention on HFC much earlier. 
    
    \begin{table}
    \centering
    \begin{tabular}{ccccc}
    \toprule 
     Acc(\%) & \multicolumn{2}{c}{LFC} & \multicolumn{2}{c}{HFC}\\
    \hline
    r  & Baseline & HARS & Baseline & HARS\\
    \hline
    4    & \textbf{15.91}  & 10.00  & \textbf{82.30} & 78.37    \\
    8    & \textbf{32.12}  & 9.99  & 40.96 & \textbf{76.25}     \\
    12   & \textbf{81.54}  & 9.54  & 15.86 & \textbf{64.20}   \\
    16   & \textbf{90.68}  & 40.00  & \textbf{11.42} & 11.06   \\
    \bottomrule
    \end{tabular}
    \caption{We report the test accuracy on baseline model and HARS model. The HARS model has a good performance on HFC, while fail to generalize on LFC.}
    \label{tab: table2}
    \vspace{-0.23cm}

    \end{table}
    
    \vspace{-0.23cm}
    
    \section{Conclusion}
    This paper investigated and explained the frequency bias in image classification. We have extended the frequency bias phenomenon with observations on feature discrimination and learning priority. Possible reasons leading to the frequency bias phenomenon are also discussed with validated hypothesis. There remains an along way to go before solving the frequency bias problem and exploiting the biased frequency towards practical model improvement. Some future works include: (1) More reasons need to be analyzed from the model perspective, e.g., whether local receptive field and layered representation benefit high-frequency feature extraction. (2) It is necessary to reaching a balance between HFC and LMFC, i.e., enhancing the biased HFC without devastating the LMFC. (3) We found in preliminary observations that the inhibited high-frequency feature is closely related to the adversarial vulnerability. It will be interesting to examine whether addressing frequency bias provides possible solution for the generalization-robustness trade-off.

\section*{Acknowledgments}

This work is supported by the National Key R\&D Program of China (Grant No.2018AAA0100604), the National Natural Science Foundation of China (Grant No.61832002, 62172094), and Beijing Natural Science Foundation (No.JQ20023).


\bibliographystyle{named}
%\bibliography{ijcai22}
\newpage

\appendix
\section{Feature Discrimination Analyses}
We report the key observations on feature discrimination in Section 3.1 and the results of KDE based inter-class variance on CIFAR-10, ResNet-50 do extend the frequency bias phenomenon. Here we provide more experimental results on different datasets and models. To investigate the feature discrimination for different frequency components, for LFC and HFC, we respectively select the frequency components with $r = {4, 8, 12, 16}$ on CIFAR-10 datasets, and with $r = {32, 64, 96, 128}$ on Restricted-ImageNet datasets (shorten as R-ImageNet). We use the outputs from the penultimate layer (512-d, ResNet-18; 2048-d, ResNet-50) to calculate the inter-class variance

\subsection*{Inter-Class Variance on CIFAR-10,ResNet-50}

\vspace{-0.15cm}

\begin{table}[ht]\label{CIFAR-R50}
    \centering
    \begin{tabular}{c|cccc}
        \toprule 
        LFC  & $X_{l}^{(4)}$ & $X_{l}^{(8)}$ & $X_{l}^{(12)}$ & $X_{l}^{(16)}$\\
        \midrule
        Acc(\%)    & 16.02  & 32.22  & 81.50 & 90.59    \\
        Variance    & 0.1897  & 0.3199  & 0.5128 & 0.5702     \\
        \midrule
        
        \midrule
        HFC  & $X_{h}^{(4)}$ & $X_{h}^{(8)}$ & $X_{h}^{(12)}$ & $X_{h}^{(16)}$\\
        \midrule
        Acc(\%)    & 82.33  & 40.81  & 15.78 & 11.4    \\
        Variance    & 0.4650  & 0.3042  & 0.1947 & 0.1485     \\
        \bottomrule
    \end{tabular}
    \caption{The test accuracy and inter-variance of different frequency components on CIFAR-10, ResNet-50}
    \label{tab:acc-variance-r50-c10}
\end{table}

\subsection*{Inter-Class Variance on CIFAR-10,ResNet-18}

\vspace{-0.15cm}

\begin{table}[ht]\label{CIFAR-R18}
    \centering
    \begin{tabular}{c|cccc}
        \toprule 
        LFC  & $X_{l}^{(4)}$ & $X_{l}^{(8)}$ & $X_{l}^{(12)}$ & $X_{l}^{(16)}$\\
        \midrule
        Acc(\%)    & 15.88  & 33.77  & 83.03 & 91.51    \\
        Variance    & 0.3583  & 0.4281  & 0.6007 & 0.5976     \\
        \midrule
        
        \midrule
        HFC  & $X_{h}^{(4)}$ & $X_{h}^{(8)}$ & $X_{h}^{(12)}$ & $X_{h}^{(16)}$\\
        \midrule
        Acc(\%)    & 84.15  & 45.44  & 23.56 & 10.17    \\
        Variance    & 0.5976  & 0.4915  & 0.3741 & 0.2701     \\
        \bottomrule
    \end{tabular}
    \caption{The test accuracy and inter-variance of different frequency components on CIFAR-10, ResNet-18}
    \label{tab:acc-variance-r18-c10}
\end{table}

\subsection*{Inter-Class Variance on R-ImageNet,ResNet-50}

\vspace{-0.15cm}

\begin{table}[ht]\label{ImageNet-R50}
    \centering
    \begin{tabular}{c|cccc}
        \toprule 
        LFC  & $X_{l}^{(32)}$ & $X_{l}^{(64)}$ & $X_{l}^{(96)}$ & $X_{l}^{(128)}$\\
        \midrule
        Acc(\%)    & 78.80  & 89.50  & 91.20 & 91.70    \\
        Variance    & 0.3626  & 0.4816  & 0.5561 & 0.5788     \\
        \midrule
        
        \midrule
        HFC  & $X_{h}^{(32)}$ & $X_{h}^{(64)}$ & $X_{h}^{(96)}$ & $X_{h}^{(128)}$\\
        \midrule
        Acc(\%)    & 14.29  & 13.2  & 11.11 & 11.11    \\
        Variance    & 0.3182  & 0.2247  & 0.1446 & 0.1447     \\
        \bottomrule
    \end{tabular}
    \caption{The test accuracy and inter-variance of different frequency components on R-ImageNet, ResNet-50}
    \label{tab:acc-variance-r50-imagenet}
\end{table}

\subsection*{Inter-Class Variance on R-ImageNet,ResNet-18}

\vspace{-0.15cm}

\begin{table}[h]\label{ImageNet-R18}
    \centering
    \begin{tabular}{c|cccc}
        \toprule 
        LFC  & $X_{l}^{(32)}$ & $X_{l}^{(64)}$ & $X_{l}^{(96)}$ & $X_{l}^{(128)}$\\
        \midrule
        Acc(\%)    & 79.00  & 90.40  & 91.20 & 91.50    \\
        Variance    & 0.5057  & 0.5775  & 0.5859 & 0.5857     \\
        \midrule
        
        \midrule
        HFC  & $X_{h}^{(32)}$ & $X_{h}^{(64)}$ & $X_{h}^{(96)}$ & $X_{h}^{(128)}$\\
        \midrule
        Acc(\%)    & 16.24  & 14.49  & 11.11 & 11.11    \\
        Variance    & 0.2988  & 0.2987  & 0.2406 & 0.1501     \\
        \bottomrule
    \end{tabular}
    \caption{The test accuracy and inter-variance of different frequency components on R-ImageNet, ResNet-18}
    \label{tab:acc-variance-r18-imagenet}
\end{table}

\noindent Table \ref{tab:acc-variance-r50-c10}-\ref{tab:acc-variance-r18-imagenet} show the results of KDE-based inter-class variance between different frequency components. All the results reflect a positive correlation between inter-variance and test accuracy, which extend the frequency bias phenomenon to a feature level. The results also demonstrate that the frequency bias phenomenon on feature discrimination is not simply an accident.

\section{Learning Priority Analyses}
\subsection{Proof of proposition}

We report the key observations on learning priority in Section 3.2. Before we visualize the learning priority by calculating the gradient map, we introduce a proposition. Here we give the proof of this proposition.

\vspace{-0.5cm}

\begin{proof}\renewcommand{\qedsymbol}{}
    Given an input image\footnote{Here we consider the image is square.} $X$, $X_{\mathcal{F}}$ is transformed Fourier image in frequency domain with same dimension as the input. Thus we have $X=\mathcal{F}^{-1}(X_{\mathcal{F}})$. We define $(i, j)$ as the coordinate in spatial domain, while $(u, v)$ as the coordinate in frequency domain. Thus, the 2D-DFT is defined as:
\end{proof}

\vspace{-1cm}

\begin{equation}
    X_{\mathcal{F}}(u,v) = \sum\limits_{i=0}^{d-1}\sum\limits_{j=0}^{d-1} X(i,j) \cdot e(i,j,u,v)
\end{equation}
where $e(i,j,u,v)$ represents $e^{-i2\pi (iu/d+jv/d)}$, $d$ is the size of input image. Similarly, the 2D-IDFT can be expressed as:

\begin{equation}
    X(i,j) = \sum\limits_{u=0}^{d-1}\sum\limits_{v=0}^{d-1} X_{\mathcal{F}}(u,v) \cdot e(u,v,i,j)
\end{equation}
we express the gradient of parameters in the spatial domain with respect to their counterparts in the frequency domain according to the chain-rule:

\begin{equation*}
    \begin{aligned}
        \dfrac{\partial{\mathcal{L}}}{\partial{X^{(k)}(i,j)}} & =\sum\limits_{u,v=0}^{d-1}             \sum\limits_{i,j=0}^{d-1} \dfrac{\partial{\mathcal{L}}}{\partial{X(i,j)}} \cdot \dfrac{\partial{X(i,j)}}{\partial{X_{\mathcal{F}}(u,v)}} \cdot \dfrac{\partial{X_{\mathcal{F}}(u,v)}}{\partial{X^{(k)}_{\mathcal{F}}(u,v)}} \\
        & \cdot \dfrac{\partial{X^{(k)}_{\mathcal{F}}(u,v)}}{\partial{X^{(k)}(u,v)}}  \\
        & =\sum\limits_{u,v=0}^{d-1} \sum\limits_{i,j=0}^{d-1} \dfrac{\partial{\mathcal{L}}}{\partial{X(i,j)}} \cdot e(i,j,u,v))\cdot 1_{(u,v) \in \mathcal{M}^{(k)}} \\
        & \cdot e(u, v, i, j)  \\
        & = \sum\limits_{u,v=0}^{d-1} (\mathcal{F}(\dfrac{\partial{\mathcal{L}}}{\partial{X}})(u,v)\cdot 1_{(u,v) \in \mathcal{M}^{(k)}}) \cdot e(u,v,i,j)\\
        & = \mathcal{F}^{-1}(\mathcal{F}(\dfrac{\partial{\mathcal{L}}}{\partial{X}}) \odot \mathcal{M}^{(k)})(u,v)\\
    \end{aligned}
\end{equation*}
Thus, we have: $\dfrac{\partial{\mathcal{L}}}{\partial{X^{(k)}}} = \mathcal{F}^{-1}(\mathcal{F}(\dfrac{\partial{\mathcal{L}}}{\partial{X}}) \odot \mathcal{M}^{(k)})$.

\vspace{0.5cm}

\subsection{Visualization of Learning Priority}
We visualize the learning priority on CIFAR-10, ResNet-50 in Fig.4, Section 3.2. Here we provide the visualization on CIFAR-10, ResNet-50,-18 and R-ImageNet, ResNet-50,-18.

\begin{figure}[htb]
% \centering
\includegraphics[scale=0.325]{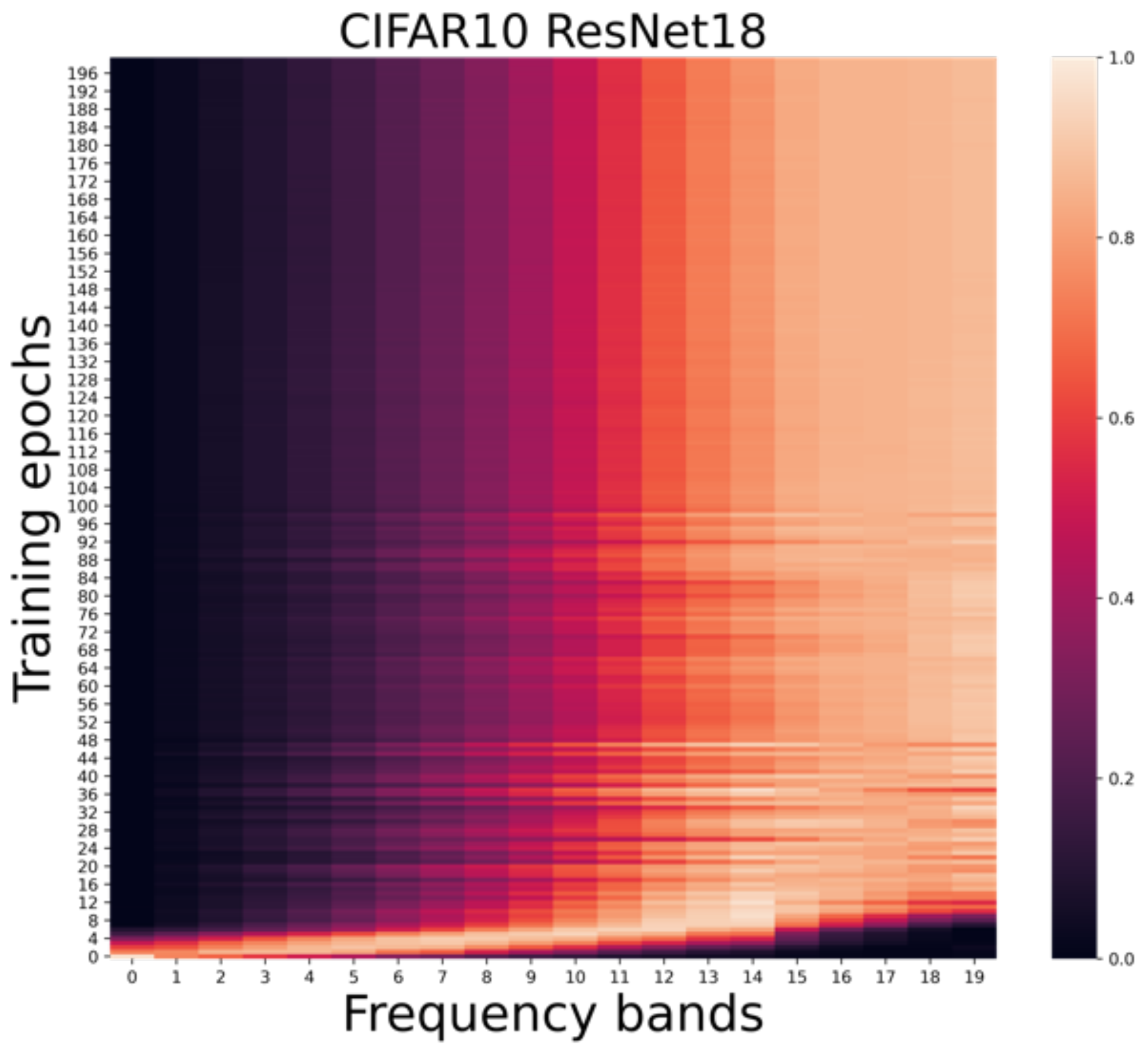}
\caption{The learning priority of baseline model (CIFAR-10, ResNet-18). It was shown in Sec 3.2.}
\label{fig:LP-CIFAR10-R18}
\end{figure}

\begin{figure}[htb]
% \centering
\includegraphics[scale=0.325]{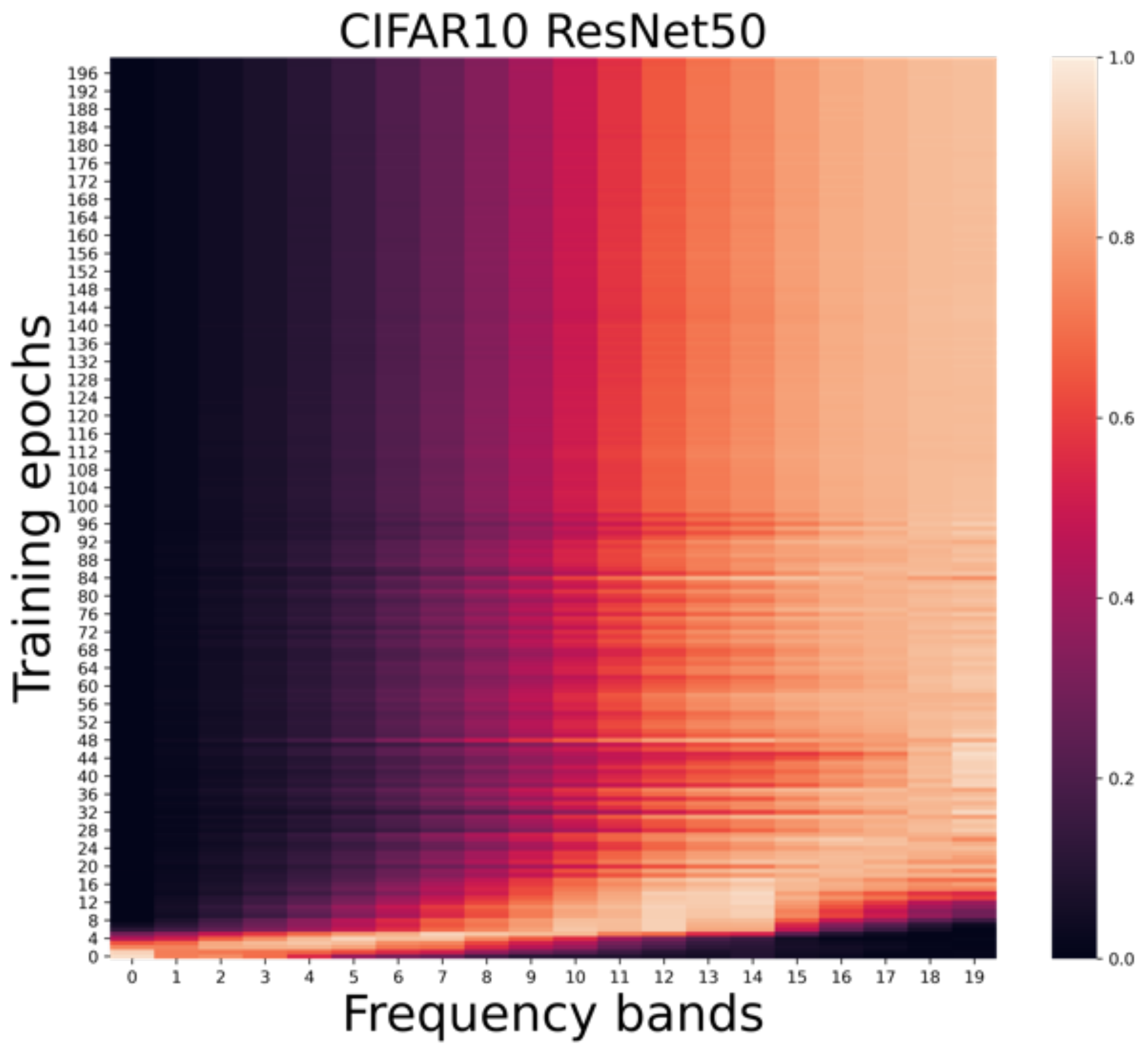}
\caption{The learning priority of baseline model (CIFAR-10, ResNet-50). It almost shows a same tendency on CIFAR-10, ResNet-18}
\label{fig:LP-CIFAR10-R50}
\end{figure}

\noindent Fig.\ref{fig:LP-CIFAR10-R50} reports the evolution of the learned frequency components (x-axis for frequency bands) during the first 200 training epochs (y-axis). The colors measure the amplitude of gradients at the corresponding frequency bands (normalized to [0, 1]). It almost shows a same tendency on CIFAR-10, ResNet-18 (Fig.\ref{fig:LP-CIFAR10-R18}). It indicates that the trends of learning priority are consistent across models.

\vspace{0.35cm}

\vspace{0.75cm}

\begin{figure}[htbp]
\centering
\includegraphics[scale=0.325]{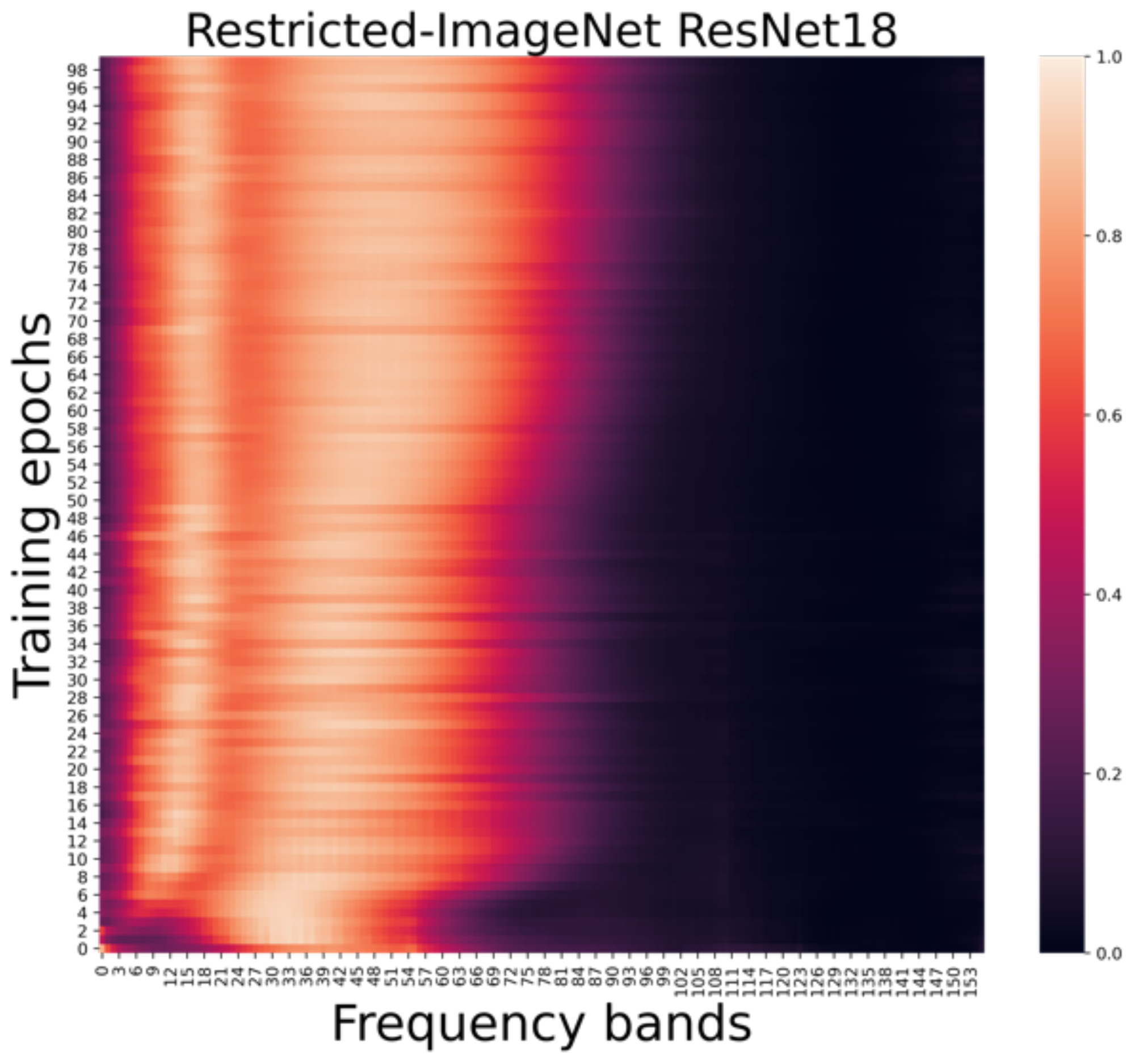}
\caption{The learning priority of baseline model (R-ImageNet, ResNet-18).}
\label{fig:LP-ImageNet-R18}
\end{figure}

\begin{figure}[htbp]
\centering
\includegraphics[scale=0.325]{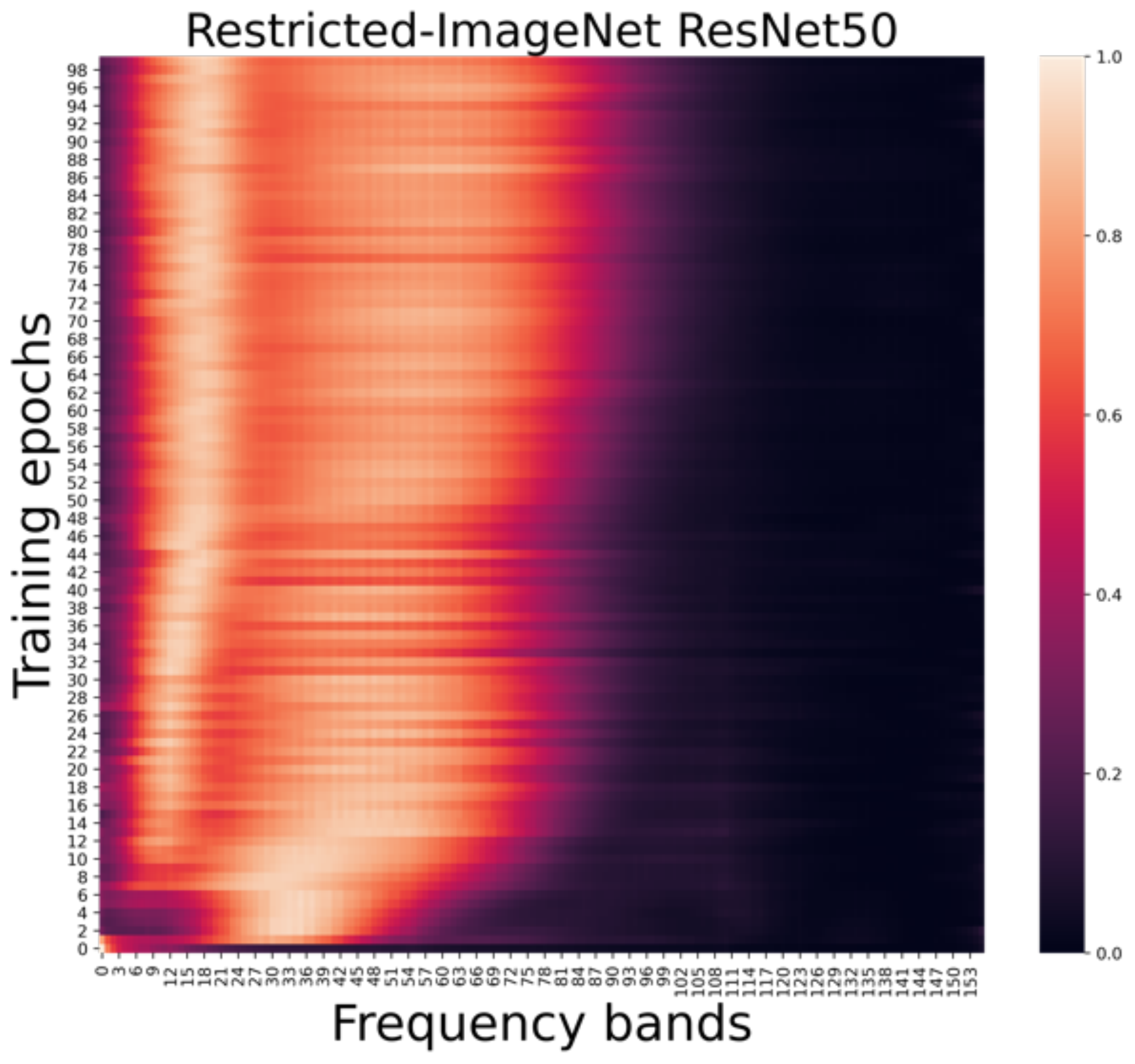}
\caption{The learning priority of baseline model (R-ImageNet, ResNet-50).}
\label{fig:LP-ImageNet-R50}
\end{figure}

\noindent Fig.\ref{fig:LP-ImageNet-R18} reports the evolution of the learned frequency components during the first 100 training epochs on R-ImageNet. We observe that the learning priority shifts from low frequency to high frequency in a quit short time and mainly focus on mid-, high-frequency components in the rest of training process. The results in Fig.\ref{fig:LP-ImageNet-R18} shows that the bias of learning priority is widespread in other datasets and does not vary by resolution. We also visualize the learning priority on ResNet-50 (shown in Fig.\ref{fig:LP-ImageNet-R50}.)

\vspace{0.35cm}

\newpage

\section{The Generalization Results of MSDA}

Here we report the generalization results of MSDA model in Section 4.2.

\begin{table}[h]
    \centering
    \begin{tabular}{c|cccccccccc}
        \toprule 
        Acc(\%) & All & $X_{l}^{(4)}$ & $X_{l}^{(8)}$ & $X_{l}^{(12)}$ & $X_{l}^{(16)}$ & \\
        \midrule
        Cutmix & \textbf{95.97}&	13.63&	31.06&	84.54&	92.27 \\
        Mixup  & 95.82	&  12.95&	\textbf{35.74}&	\textbf{85.37}&	\textbf{92.76}\\
        Baseline  & 95.42 &	\textbf{15.91}&	32.12&	81.54&	90.68 \\
        \bottomrule
    \end{tabular}
    \caption{The test accuracy on low-frequency images CIFAR-10, ResNet-50}
    \label{tab:acc-variance}
\end{table}

\begin{table}[h]
    \centering
    \begin{tabular}{c|cccccccccc}
        \toprule 
        Acc(\%) & All & $X_{h}^{(4)}$ & $X_{h}^{(8)}$ & $X_{h}^{(12)}$ & $X_{h}^{(16)}$ & \\
        \midrule
        Cutmix  &\textbf{95.97} &81.63 & \textbf{47.34} &\textbf{17.99} & 10.87\\
        
        Mixup &95.82 &	\textbf{82.63}&	42.43&	13.41&	10.07 \\
        
        Baseline  & 95.42 &	82.3 &	40.96&	15.86&	\textbf{11.42} \\
        \bottomrule
    \end{tabular}
    \caption{The test accuracy on high-frequency images CIFAR-10, ResNet-50}
    \label{tab:acc-variance}
\end{table}

\end{document}